\documentclass[nohyperref]{article}

\usepackage{microtype}
\usepackage{graphicx}
\usepackage{subfigure}
\usepackage{booktabs} 

\usepackage{float}
\usepackage{hyperref}
\usepackage{multirow}
\usepackage{algorithm}
\usepackage{algpseudocode}

\usepackage[accepted]{icml2022}

\usepackage{amsmath}
\usepackage{amssymb}
\usepackage{mathtools}
\usepackage{amsthm}

\usepackage[capitalize,noabbrev]{cleveref}

\theoremstyle{plain}

\theoremstyle{definition}

\theoremstyle{remark}

\usepackage[textsize=tiny]{todonotes}
\usepackage{colortbl}
\usepackage{comment}
\usepackage{verbatim}
\icmltitlerunning{BiPOCO: Bi-directional Trajectory Prediction with Pose Constraints for Pedestrian Anomaly Detection}

\begin{document}

\twocolumn[
\icmltitle{BiPOCO: Bi-directional Trajectory Prediction with Pose Constraints for Pedestrian Anomaly Detection}



\icmlsetsymbol{equal}{*}

\begin{icmlauthorlist}
\icmlauthor{Asiegbu Miracle Kanu-Asiegbu}{ME}
\icmlauthor{Ram Vasudevan}{ME}
\icmlauthor{Xiaoxiao Du}{NAMES}
\end{icmlauthorlist}

\icmlaffiliation{ME}{Department of Mechanical Engineering, University of Michigan, Ann Arbor, MI, USA}
\icmlaffiliation{NAMES}{Department of Naval Architecture and Marine Engineering, University of Michigan, Ann Arbor, MI, USA}

\icmlcorrespondingauthor{Asiegbu Miracle Kanu-Asiegbu}{akanu@umich.edu}
\icmlcorrespondingauthor{Ram Vasudevan}{ramv@umich.edu}
\icmlcorrespondingauthor{Xiaoxiao Du}{xiaodu@umich.edu}

\icmlkeywords{Deep Learning, Trajectory Prediction, Human Pose, Anomaly Detection}

\vskip 0.3in
]



\printAffiliationsAndNotice{}  

\setlength{\abovedisplayskip}{3pt}
\setlength{\belowdisplayskip}{3pt}

\setlength{\belowcaptionskip}{-10pt}
\setlength{\textfloatsep}{1\baselineskip plus 0.2\baselineskip minus 0.5\baselineskip}
\setlength{\intextsep}{10pt plus 2pt minus 2pt}

\begin{abstract}
We present BiPOCO, a Bi-directional trajectory predictor with POse COnstraints, for detecting anomalous activities of pedestrians in videos. In contrast to prior work based on feature reconstruction, our work identifies pedestrian anomalous events by forecasting their future trajectories and comparing the predictions with their expectations. We introduce a set of novel compositional pose-based losses with our predictor and leverage prediction errors of each body joint for pedestrian anomaly detection. Experimental results show that our BiPOCO approach can detect pedestrian anomalous activities with a high detection rate (up to 87.0\%) and incorporating pose constraints helps distinguish normal and anomalous poses in prediction. This work extends current literature of using prediction-based methods for anomaly detection and can benefit safety-critical applications such as autonomous driving and surveillance. Code is available at \url{https://github.com/akanuasiegbu/BiPOCO}.
\end{abstract}

\vspace{-8mm}

\section{Introduction}
\label{introduction}
Pedestrian anomaly detection (PAD) refers to the problem of identifying pedestrian activities and events that do not conform to expected behavior (i.e., ``anomalous'') from video sequences \cite{li2013anomaly,andrews2016transfer, deecke2021transfer}. This is an essential task in safe vehicle autonomy. If a pedestrian suddenly breaks into a run or jumps in front of the vehicle, or if a child throws a ball into the street, an intelligent autonomous vehicle must be able to recognize these activities as potentially anomalous, so that appropriate path planning and control strategies may be used to avoid collision and ensure safety. 

\vspace{-1mm}

The PAD problem is set up with a training set and a testing set, where the training videos contain only “normal” pedestrian activities (e.g., walking), and the testing videos can contain both normal
and anomalous activities (e.g., running, jumping, throwing a bag, etc.). The goal of PAD is to learn the normal pedestrian activity patterns present in the training data and to detect the
anomalies in the testing set.

\vspace{-1mm}

Anomalous events typically occur far less frequently than normal activities \cite{sultani2018real} and it is usually difficult to obtain training data that contains large quantities of labeled anomalous events. Thus, existing solutions to the PAD problem are predominantly  unsupervised learning approaches that learn feature representations for normal activities given training data that only contains normal instances. Such techniques include sparse coding \cite{cewu2013}, Minimum Volume-set estimation \cite{thomas2016learning, thomas2017anomaly}, and topic modeling \cite{isupova2016anomaly, girdhar2016anomaly}. With the rise of deep learning approaches, encoder-decoder-style architectures have been developed for PAD using long short-term memory (LSTM) units \cite{srivastava2015unsupervised}, autoencoders \cite{hasan2016learning,chong2017abnormal, luo2017revisit}, convolutional neural networks (CNNs) \cite{nguyen2019anomaly}, and generative adversarial networks (GANs)  \cite{ravanbakhsh2017abnormal, chen2021nm}. In these works, feature representations for ``normal'' activities are typically learned by reconstructing training video frames, and anomalies are detected by computing reconstruction errors for the test instances (higher reconstruction error indicates anomalous events). Such pixel-based
features learned through reconstruction can be unstructured and high-dimensional, which can result in sensitivity to noise, lack of interpretability, and redundancy \cite{morais2019learning}.

\begin{figure*}[htp!]
\begin{minipage}{0.2\textwidth}
   \centering
  \includegraphics[width=\textwidth]{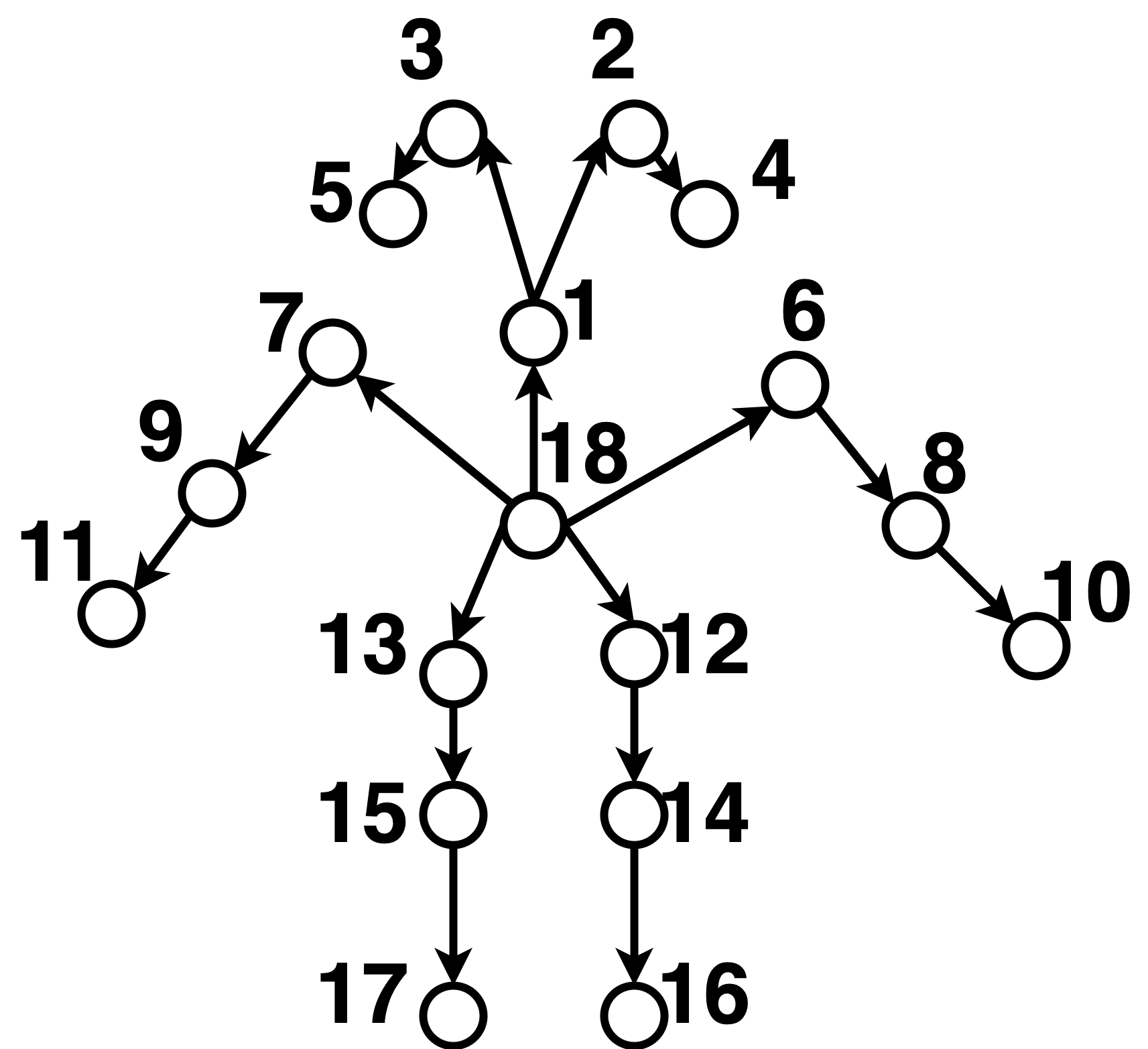}
  \caption{Skeleton Pose.}
 \label{fig:sub1}
  \end{minipage}\qquad
  \begin{minipage}{0.78\textwidth}
    \centering
  \includegraphics[width=\textwidth,trim=0 0cm 0 0,clip]{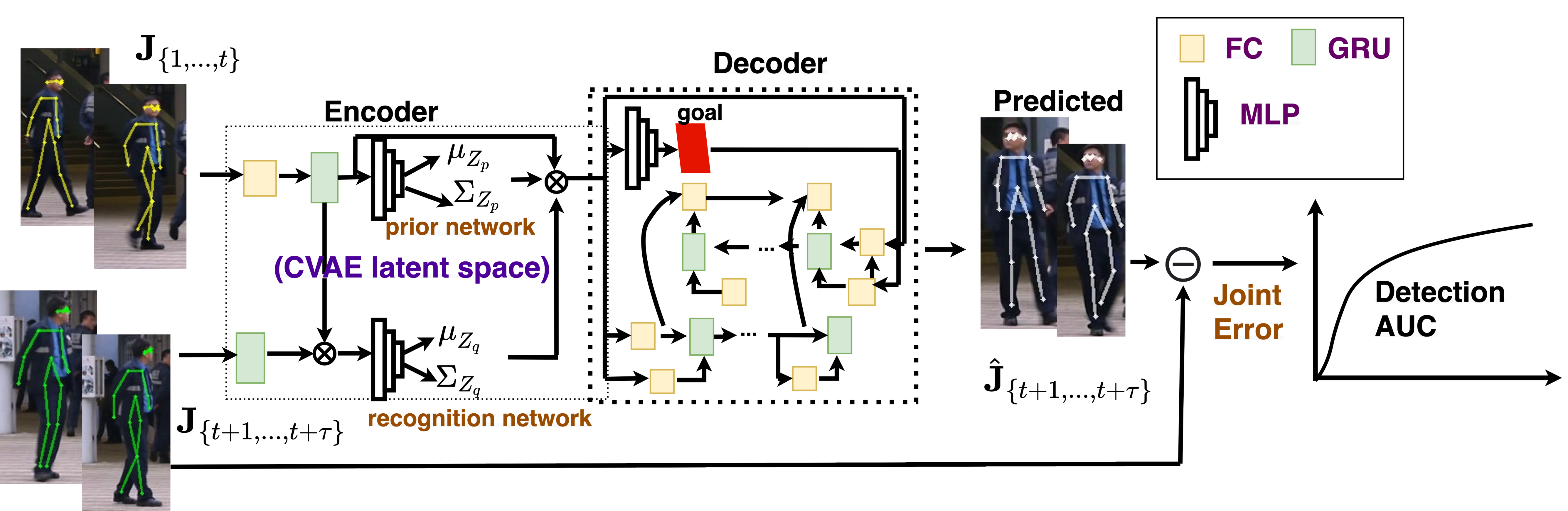}
  \vspace{-10mm}
  \caption{BiPOCO pipeline diagram.}
  \label{fig:sub2}
  \end{minipage}
\label{fig:diagram}
\end{figure*}


\vspace{-1mm}
In this work, instead of relying on reconstruction error, we propose to identify anomalous events by evaluating their expectation and introducing a  prediction-based pipeline for pedestrian video anomaly detection. Our main contributions include:
\begin{itemize}
\vspace{-4mm}
\item \textbf{BiPOCO}, a novel bi-directional  trajectory predictor that predicts  pedestrian 2-D skeleton pose trajectories while incorporating a novel set of compositional pose constraints based on the physical structure of human skeletons (poses). 
\vspace{-2mm}
\item An anomaly detection pipeline based on BiPOCO joint prediction errors to distinguish anomalous pedestrian events from normal instances. Our prediction-based pipeline outperformed previous reconstruction-based methods. We introduce pose-based losses, which resulted in increased detection accuracy of 11.4\%-14.8\% compared with prior BiTrap trajectory predictors without pose-based losses.
\vspace{-2mm}
\item Extensive ablation studies across pose-based loss terms on two benchmark datasets. We observed that constraining the locations of the extremities (the joint endpoints) has significant impact on improving both joint prediction and anomaly detection results, while a combination of bone (limbs) and joint-based losses also contribute to improved detection results. 
\end{itemize}
\vspace{-4mm}
This work offers a solution to the PAD problem by leveraging recent advances in trajectory forecasting techniques, and provides insights for incorporating structure-aware pose constraints.





\vspace{-3mm}

\section{Our Approach}
\label{sec:method}
\vspace{-2mm}
BiPOCO takes pedestrian skeleton pose trajectories from video data as input and outputs predicted poses (sets of joint locations) for all pedestrians at future timesteps. Section \ref{sec:bitrap} describes the foundation of BiPOCO, a CVAE-based bi-directional trajectory predictor. Section \ref{sec:posecons} presents three novel compositional pose constraints that incorporate  skeleton structure when training BiPOCO. Section~\ref{sec:anomaly} describes the anomaly detection mechanism, where frame-level anomaly  scores can be computed based on joint errors after obtaining BiPOCO predictions to detect anomalous pedestrian events. 

\vspace{-3mm}

\subsection{Bi-directional Trajectory Predictor}
\label{sec:bitrap}
\vspace{-2mm}
Our bi-directional trajectory predictor is built upon the \textit{BiTraP} model \cite{yao2021bitrap}. In BiTraP,  input sequences are encoded using a gated-recurrent unit (GRU) encoder network to obtain encoded feature vectors for input trajectory sequences. Then, a conditional variational autoencoder (CVAE) is used to predict a Gaussian distribution mean and covariance from input pedestrian trajectory sequences at both observed and ground truth future timesteps. Next, a latent variable $Z$ is sampled from the learned distribution and concatenated with the encoded feature vectors to predict a goal location for the trajectory. The predicted goal is used in a bi-directional trajectory generation network, where a set of fully connected (FC) networks and GRUs are used to predict future trajectories in both forward (from current timestep to goal location at the prediction horizon) and backward (from predicted goal to the current timestep) directions. Figure~\ref{fig:sub2} shows the diagram for the network architecture. 


\vspace{-1mm}
The standard BiTraP model only predicts bounding box locations. To predict the full body model of the dynamic skeleton features of pedestrians, we modified the network to predict COCO-style skeleton joint (``keypoint'') locations (see Figure~\ref{fig:sub1} for the location of the 17 keypoints). Note that we appended \#18, a root joint, as the mean of left and right shoulder and hip joints. We use AlphaPose \cite{fang2017rmpe, xiu2018poseflow} to generate 2-D skeleton joint locations for all pedestrian sequences. During training, given joint locations of pose sequence length $\tau+\delta$ (where $\tau$ and $\delta$ are observation and prediction horizons, respectively), a bi-directional trajectory prediction loss $L_T$ is used to minimize the L2 error between the predicted goal and poses as well as the Kullback–Leibler divergence  between prior and recognition networks in the CVAE  (see Appendix \ref{appendix:cvae}).





%

\vspace{-3mm}
\subsection{Compositional Pose Constraints}
\label{sec:posecons}
\vspace{-2mm}
We introduce a novel set of  compositional pose constraints (based on bone, joints, and endpoints) on top of the bi-directional trajectory predictor loss to take advantage of the physical properties of human skeleton structure. 

\vspace{-1mm}
%

\textbf{Bone Loss}: Following \cite{sun2017compositional}, we define a bone as a directed vector pointing from a joint to its parent, $\mathbf{B}_{n} = {\mathbf{J}}_{parent(n)} -{\mathbf{J}}_{n}$, where the function $parent(n)$ returns the index of parent joint for the $n^{th}$ joint $\mathbf{J}_n$.  The bone-based loss is written as $L_B = \sum_{n=1}^{18} \left \| \hat{\mathbf{B}}_{n} - \mathbf{B}_{n} \right \|$,
%
where $\left\|\cdot \right \|$ denotes L1-norm, ${\mathbf{B}}$ is the ground truth bone vectors as obtained by AlphaPose, and  $\hat{\mathbf{B}}$ is the bone vectors computed from the predicted joints by our BiPOCO predictor. The bone loss incorporates geometric structure by accounting for the shapes (directions and lengths) of limbs.

\vspace{-1mm}

\textbf{Endpoint Loss}: 
Human extremities carry powerful information of the behavior and action of the person and can be regarded as a compact semantic representation of the
human posture \cite{yu2009human}. However, the locations of the extremities (the joint endpoints) are often poorly reconstructed and/or predicted due to their wide range of motion and noises caused by self occlusion \cite{huang2018deep, du2020unsupervised}. We propose to incorporate the errors between the predicted endpoint joints and their parent joint locations, so that the algorithm may better learn the motion of the pedestrians' extremities and enhance the overall pose prediction. We consider six endpoint tracks - left and right arms, legs, and faces. We write out the endpoint loss equation for the left arm (LA) below as an example
%
\begin{align}
\label{eqn:altendpoint1}
\small
\begin{split}
\mathcal{L}_{LA} &= \left \| \sum_n (\hat{\mathbf{J}}_n - \mathbf{J}_{parent(n)})\right \|,  n\in \left \{ 6,8,10\right \}
\\
&=  \left \| \left (\hat{\mathbf{J}}_{10} - {\mathbf{J}}_{8} \right ) + \left (\hat{\mathbf{J}}_{8} - {\mathbf{J}}_{6} \right ) + \left (\hat{\mathbf{J}}_{6} - \mathbf{J}_{18} \right ) \right \| ,
\end{split}
\end{align}
%
where $n$ denotes the joint endpoints along the left arm,  the indices $\left \{ 6,8,10\right \}$ correspond to left shoulder, elbow, and wrist, respectively, and 18 is the root joint. Similarly, we can write out loss terms for the right arm (RA), left face (LF), right face (RF), left leg (LL) and right leg (RL) (see Appendix \ref{appendix:endpoint_loss_equations}). The endpoint loss will be computed as 
\begin{equation}
\label{eqn:endpointloss}
L_{E} = \mathcal{L}_{LA} + \mathcal{L}_{RA} + \mathcal{L}_{LF} + \mathcal{L}_{RF} + \mathcal{L}_{LL} + \mathcal{L}_{RL}.
\end{equation}

\vspace{-1mm}

\textbf{Joint Loss}: The trajectory loss in Section \ref{sec:bitrap} was computed using normalized, relative pose coordinates. To account for the skeleton’s absolute position in relation to the entire scene in the image frame, we also include a global joint loss term, where the difference between the ground truth and predicted joints in a global (image) coordinate frame was computed during training, $L_J = \sum_{n=1}^{18} \left \| \hat{\mathbf{J}}_{n} - \mathbf{J}_{n} \right \|$,
%
%
where $\hat{\mathbf{J}}_{n}$ and ${\mathbf{J}}_{n}$ are the predicted and ground truth joint locations. 

The overall training loss for BiPOCO is
\begin{equation}
\label{eqn:allloss}
L = L_{T} + \alpha L_{B} + \beta L_{E} + \gamma L_{J},
\end{equation}
where $\alpha$, $\beta$ and $\gamma$ are weights associated with each of the loss components. Currently $\alpha$, $\beta$, and $\gamma$ are set to one. We will present an ablation study on various combinations of the loss components in the experiments. 

\vspace{-3mm}

\subsection{Anomaly Event Detection}
\label{sec:anomaly}
\vspace{-2mm}
BiPOCO outputs predicted skeleton joint locations for each pedestrian in each frame. Our hypothesis is that the predicted poses of pedestrians performing anomalous activities will deviate further from the ground truth trajectory compared to normal activities, since the BiPOCO predictor was trained to learn normal (walking) motion from the training data. Thus, the prediction error between the predicted and ground truth pedestrian poses can be used as an indicator for identifying anomalous events. We first compute person-level joint errors via a weighted  squared error $e^p_{t} = \sum_{k=1}^{17} w_{kt} (J_{kt} - \hat{J}_{kt})^2$,
%
%
where $J_{kt}$ is the ground truth (obtained via AlphaPose) skeleton joint location for $k^{th}$ joint at timestep $t$, $\hat{J}_{kt}$ is the BiPOCO prediction, and  $w_{kt}$ is the confidence score obtained by AlphaPose to indicate the visibility of the $k^{th}$ joint  at timestep $t$. 

\vspace{-1mm}
Since we use a sliding window method to handle pedestrian pose sequences as inputs and outputs for BiPOCO, there can exist multiple predictions for the same pedestrian instance. In this case, we follow \cite{kanu2021leveraging} and use two error measures, named ``Summed Error'' and ``Flattened Error'', to gather anomaly scores for the same pedestrian across multiple predictions. The Summed Error sums up the skeleton joint errors for all prediction timesteps in each sequence, and the Flattened Error averages prediction errors at the same timestep across pedestrian sequences.
\vspace{-1mm}

After computing the skeleton joint errors for each pedestrian $i$ at each frame (timestep) $t$, we aggregate the skeleton errors into a frame-level anomaly score. The frame-level anomaly score at timestep $t$ is computed by max pooling the weighted squared error, $e^p_{t}$, of all skeleton instances in that frame as $ e^f_t = \max_{\forall p\in \mathcal{P}_t} e^p_{t}$,
%
%
where $\mathcal{P}_t$ refers to the set of all pedestrian instances appearing in the frame $t$. We use max pooling to suppress the influence of normal instances present in the
scene, since normal activities correspond to smaller prediction errors and the number of normal instances can vary largely in real videos  \cite{morais2019learning}. The frame-level
anomaly scores for all frames $t$ are the final outputs of our
pipeline for anomaly detection.



\vspace{-3mm}

\section{Experiments}
\label{sec:experiments}
\vspace{-2mm}
We examine the performance of our BiPOCO predictor on two benchmark pedestrian anomaly detection datasets, \textit{Avenue} \cite{cewu2013} and \textit{ShanghaiTech (ST)} \cite{luo2017revisit} datasets. The Avenue dataset contains 16 training videos and 21 testing videos ($640\times360$ pixels) collected at 25 frames per second from a fixed location (15,328 training frames and 15,324 testing frames). The ShangahiTech dataset includes 330 training videos and 107 testing videos ($856\times480$ pixels) collected from 13 locations with different camera angles. In both datasets, pedestrian walking activities are labeled as ``normal''. Events such as riding a bike/skateboard, loitering, fighting, running, jumping and throwing objects are labeled as anomalous events.  We also report results on HR-Avenue and HR-ShanghaiTech, an abbreviated version of both datasets that ignored video frames that contain  non-detectable poses or non-human anomalies \cite{morais2019learning}. Frame-level ROC (receiver operating characteristic)  AUC (area under curve) is used as the benchmark
evaluation metric to evaluate the anomaly detection results following \cite{hasan2016learning, morais2019learning}.

\vspace{-1mm}

We report the AUC detection results of our proposed BiPOCO approach with comparison methods (see Table~\ref{table:auc_compare_av_st}). BiPOCO outperforms  previous reconstruction-based approaches, the Conv-AE \cite{hasan2016learning} and TSC sRNN \cite{luo2017revisit} on the HR-Avenue and ShanghaiTech datasets. BiPOCO also performs  better than a previous frame-based prediction approach using U-Net \cite{liu2018future} that generates entire future frames instead of leveraging human trajectories. BiTraP \cite{kanu2021leveraging} and MPED-RNN \cite{morais2019learning} both consider human trajectories, but BiTraP only predicts bounding boxes while MPED-RNN considers human skeletons but does not include any of our structured pose-based loss terms. Our BiPOCO achieves $\sim$8\% improvement in detection AUC on the HR-Avenue dataset, which shows the effectiveness of including both full skeleton representation  and the pose constraints. BiPOCO  is currently among the leading methods on HR-Avenue and ShanghaiTech and outperforms previous frame and skeleton prediction-based methods, but it has not yet surpassed \cite{rodrigues2020multi}, which considers trajectories across multiple timescales (Current BiPOCO is trained at a fixed timescale - we elaborate on this next).


\vspace{-5mm}

\begin{table}[h!]
\centering
\caption{AUC comparison results with existing methods.}
\label{table:auc_compare_av_st}
\resizebox{.48\textwidth}{!}{
\begin{tabular}{c|c|c|c|c} 
 \hline
  & \textit{Avenue} &  \textit{HR-Av.} & \textit{ST} & \textit{HR-ST}\\ 
 \hline

Stacked LSTM (baseline) & 0.631 & - & 0.655 & -\\ 
Conv-AE \yrcite{hasan2016learning} & 0.702 & 0.848 & 0.704  & 0.698 \\ 
TSC sRNN \yrcite{luo2017revisit} & 0.817  & - & 0.680  & - \\
U-Net \yrcite{liu2018future}  & 0.849  & 0.862 &   0.728  & 0.727\\
MPED-RNN \yrcite{morais2019learning}& 0.863  & 0.863 & 0.734 & 0.754\\
\textit{Multi-Timescale} \yrcite{rodrigues2020multi} & \textit{0.828}  & \textit{0.883} & \textit{0.760} & \textit{0.770}\\
BiTraP bbox only \yrcite{kanu2021leveraging}  & 0.720  & - & 0.719 & - \\
\textbf{BiPOCO (Ours)} & 0.802 & 0.870 & 0.737 & 0.749 \\

 \hline
\end{tabular}}
\end{table}

We conducted an ablation study on the compositional pose constraints  on both Avenue and HR-Avenue datasets (see Table~\ref{table:frame_auc_error_all}). The observed and predicted pedestrian sequence lengths $\tau=\delta=\{3,5,13,25\}$ correspond to activities  at varying timescales. For example, ``jumping'' can occur at a very short timescale whereas ``loitering'' may be
a long-term anomaly \cite{rodrigues2020multi}. We also report anomaly detection results using both Summed Error (SE) and Flattened Error (FE) measures (see Section \ref{sec:anomaly}). As shown, adding the endpoint loss term achieves the highest overall AUC in both Avenue and HR-Avenue, outperforming trajectory loss only (None) without any pose constraints. Using the Flattened Error measure in general outperforms the Summed Error measure, which indicates that it is advantageous to average across the multiple predictions at the same timestep generated by the sliding window sequences. The endpoint loss achieves a high detection performance at shorter timescales (3), but deteriorates as the timescale increases, possibly because (i) the pedestrian extremities start to have greater variations in longer sequences and (ii) there are fewer longer sequences for training with complete endpoints, as noises and self-occlusion can lower the AlphaPose confidence on the extremities. The combination of bone and joint losses also achieves a high AUC, particularly in longer timescales (13 and 25), which shows that constraining the physical structure (limb length, shape, global joints) of the pedestrians helps better distinguish the predicted poses for normal and anomalous activities. We provide in Appendix \ref{appendix:joint_error}
 an additional analysis on joint error results and find that the endpoint loss and the combination of Bone-Endpoint and Bone-Joint losses produces a larger margin between the predicted joint errors of normal and anomalous instances, which  contributes to improved detection performance. 

\vspace{-4mm}

\begin{table}[h!]
\caption{Ablation study of the pose constraints: Bone(B), Endpoint(E), Joint(J), All losses(All), and Trajectory loss only(None). \textcolor{blue}{Blue}: best AUC at each timescale. \textbf{Bold}: best AUC overall.}
\label{table:frame_auc_error_all} 
\centering
\resizebox{.5\textwidth}{!}{
\begin{tabular}{p{1cm}|c|cccccccc} 
 \hline
     \multirow{2}{1cm}{\textbf{Dataset}}& \multirow{2}{*}{$\mathcal{T}$}& \multicolumn{8}{c}{\textbf{Loss Terms}} \\
 \multirow{4}{1.2cm}{ \hspace{2mm} HR-Av. - SE}&   & B & E &J & B-E& B-J& E-J& All&None\\ \hline
  & 3 & 0.715 & \textcolor{blue}{0.852}&0.735&0.830&0.697&0.722&0.689 &0.757 \\
   & 5& 0.725	&	0.779		&0.733	&	\textcolor{blue}{0.784}	&	0.754	&	0.714	&	0.729 & 0.748		\\
   & 13& 0.754	&	0.748 &		\textcolor{blue}{0.779}	&	0.777 &		0.762		& 0.722&		0.707 & 0.753\\
   & 25& 0.663	&	0.696 &		0.670&		0.661	&	\textcolor{blue}{0.708}&		0.664&	0.670 & 0.668	\\\hline
  \multirow{4}{1.2cm}{ \hspace{2mm} HR-Av. - FE}&  3&0.712 &		\textbf{\textcolor{blue}{0.870}}	&	0.725	&	0.826&		0.708	&	0.711&0.685 &0.758\\					
   & 5 &0.721	&	\textcolor{blue}{0.788}	&	0.731&		0.769	&	0.739&		0.714	&	0.730 & 0.752\\
    & 13 & \textcolor{blue}{0.779}	&	0.745	&	0.750	&	0.759		&0.756	&	0.729	&	0.767 & 0.768\\
    &25 & 0.670	&	0.717	&	0.692	&	0.690	&	\textcolor{blue}{0.748}	&	0.678	&	0.720 & 0.698	\\	\hline
    \multirow{4}{1.1cm}{ \hspace{2mm} Avenue - SE}& 3 & 0.678	&	\textcolor{blue}{0.772}	&	0.698	&	0.745	&	0.677	&	0.683	&	0.657 & 0.707	\\
   & 5     & 0.693	&	\textcolor{blue}{0.740}	&	0.701	&	0.734	&	0.725	&	0.688	&	0.685 & 0.704 \\
   & 13 & 0.704	&	0.694	&	0.702	&	0.727	&	\textcolor{blue}{0.723}	&	0.676	&	0.653  & 0.709\\
   & 25 & 0.634	&	0.669	&	0.645	&	0.640	&	\textcolor{blue}{0.677}	&	0.638	&	0.643 & 0.635 \\\hline 
     \multirow{4}{1.1cm}{ \hspace{2mm} Avenue - FE}& 3 & 0.690	&	\textbf{\textcolor{blue}{0.802}}	&	0.695	&	0.761	&	0.693	&	0.693	&	0.667 & 0.720\\
   & 5 &	0.727	&	\textcolor{blue}{0.786}	&	0.744	&	0.781	&	0.766	&	0.741	&	0.720 & 0.747\\
   & 13 &	0.731	&	0.710	&	0.718	&	\textcolor{blue}{0.752}	&	\textcolor{blue}{0.752}	&	0.711	&	0.702 & 0.727 \\
   & 25 &	0.640	&	0.692	&	0.654	&	0.667	&	\textcolor{blue}{0.705}	&	0.673	&	0.689 & 0.691\\ \hline
  \end{tabular}}
\end{table}

\vspace{-4mm}

\section{Limitations}
\vspace{-1mm}
Our approach shows encouraging results leveraging pose prediction and trajectory forecasting techniques for pedestrian anomaly detection. Adding novel compositional pose constraints helps distinguish normal and anomalous poses and thus, can improve anomaly detection performance. 
\vspace{-1mm}

This approach can be potentially applied for vision-only self-driving, as the pipeline only requires camera footage and does not require
more advanced sensors such as LiDARs (Light Detection and Ranging), etc. However, current results are tested on stationary camera videos only and it would be worthwhile to further evaluate the approach on naturalistic driving datasets. Additionally, the current pipeline only warns about the occurrence of anomalous events and does not distinguish specific activity types; future work can include leveraging the predicted pose information to provide additional information about the motions and behaviors of pedestrians. 
Furthermore, our approach does not require manual annotation of skeletons a priori, but relies on the accuracy of image-based pose estimators. Future work on improving the accuracy/reducing noise on pose estimation (e.g., using temporal information of the pose sequence instead of frame-based estimation) can be investigated.

\vspace{-4mm}
\section{Conclusion}
\vspace{-2mm}
We present BiPOCO, a bi-directional trajectory predictor with pose constraints for pedestrian anomaly detection.  We achieve high  detection AUC (87\% on HR-Avenue) while demonstrating the effectiveness of adding endpoint, bone and joint-based losses for incorporating skeleton structure.






\clearpage



\section*{Acknowledgment}
This work was supported by a grant from Ford Motor Company via the Ford-UM Alliance under award N028603. This material is based upon work supported by the Federal Highway Administration under contract number 693JJ319000009. Any options, findings, and conclusions or recommendations expressed in the this publication are those of the author(s) and do not necessarily reflect the views of the Federal Highway Administration.

\bibliography{citations}
\bibliographystyle{icml2022}

\newpage
\appendix
\onecolumn



\section*{Appendix}

\section{CVAE Details and the  Trajectory Loss Equation} 
\label{appendix:cvae}


We present additional details on the conditional variational autoencoder (CVAE) in the bi-directional trajectory predictor in BiPOCO. The CVAE learner can be divided into two parts, a prior network and a recognition network \cite{sohn2015learning}, as shown in Figure~\ref{fig:sub2}. The prior network predicts the distribution mean and covariance $\mathcal{N}(\mu_{Z_p}, \Sigma_{Z_p})$ of trajectories in observed timesteps only, while the recognition network predicts the distribution mean and covariance $\mathcal{N}(\mu_{Z_q},\Sigma_{Z_q})$ of both observed and ground truth (target) future trajectories. Both distributions are assumed to follow a Gaussian distribution. A Kullback–Leibler divergence ($KLD$) loss between $\mathcal{N}(\mu_{Z_p}, \Sigma_{Z_p})$ and $\mathcal{N}(\mu_{Z_q},\Sigma_{Z_q})$ is optimized so that the dependency between observed and ground truth (target) trajectory sequences at future timesteps may be captured.  A latent variable $Z$ is sampled from $\mathcal{N}(\mu_{Z_q},\Sigma_{Z_q})$ and concatenated with the input encoder hidden state to predict the pedestrian's future goal pose $\hat{G}_t$ using a 3-layer multi-layer perceptron  (MLP) network.

As shown in Figure~\ref{fig:sub1}, the human pose is structured as a tree.  Limbs can be expressed using joints in a parent-child relationship (for example, the left shoulder joint 6 is the parent for the left elbow joint 8, and left wrist joint 10 is the child of the left elbow joint 8). The arrows indicate such joint parent-child relationships. Following \cite{sun2017compositional}, we define a bone as a directed vector pointing from a joint to its parent, $\mathbf{B}_{n} = {\mathbf{J}}_{parent(n)} -{\mathbf{J}}_{n}$, where the function $parent(n)$ returns the index of parent joint for the $n^{th}$ joint $\mathbf{J}_n$. The root joint, 18, is regarded as the parent for the shoulders (6, 7), nose (1), and hip joints (12, 13). The parent of the root joint is the image origin. Then, the position of each joint relative to its parent joint (essentially, the ``bone'' vectors) can be computed. These relative joint coordinates were used as inputs to our BiPOCO model, so as to constrain the range of joint location values while  preserving the local skeleton joint relationships.

The bi-directional trajectory predictor also includes a bi-directional decoder \cite{yao2021bitrap}, which contains both forward and backward recurrent neural networks (RNNs) composed of a sequence of gated-recurrent units (GRUs). The forward RNN is similar to a regular RNN decoder except its output is not transformed to trajectory space. The backward RNN is first initialized from the input encoder hidden state. Then, the backward RNN takes the estimated goal as input and propagates backwards from time $t+\delta$ to $t+1$, so that backward hidden state is updated from the goal to the current location. Here, the ``goal''  is defined as the pedestrian's goal pose at the last timestep of the prediction sequences. The goal vectors (relative coordinates) at the last timestep can be written as $\hat{G}_{t} = \hat{\mathbf{B}}_{t+\delta}$, where $t$ is the current timestep and $\delta$ is the prediction horizon.
 Both forward and backward hidden states for the same timestep are then concatenated to predict the final pose at that timestep. The trajectory loss function is a combination of the L2 loss  for the goal pose (in relative coordinates), the L2 loss for the entire trajectory, and the KL-divergence loss between
prior and recognition networks, written as   
\begin{equation}
\label{eqn:trajectoryloss}
L_{T} = \left \| {\mathbf{B}}_*^{t+\tau} -  \hat{\mathbf{B}}_*^{t+\tau} \right \|_2 + \sum_{k=t+1}^{t+\tau} \left \| {\mathbf{B}_*}^{k} -  \hat{\mathbf{B}}_*^{k} \right \|_2 + D_{KL}(P||Q),
\end{equation}
where $\mathbf{B}_*^{t}$ denotes the ground truth pedestrian skeleton keypoints in relative coordinates (i.e., the bone vectors) at timestep t,  $\hat{\mathbf{B}}_*^{t}$ denotes the predicted pedestrian skeleton bones at timestep $t$, $\mathbf{B}_*^{t+\tau}$ and $\hat{\mathbf{B}}_*^{t+\tau}$ is the ground truth and predicted final goal pose of the pedestrian at the last timestep of the prediction horizon, and $D_{KL}(P||Q)$ denotes the KL-Divergence between $\mathcal{N}(\mu_{Z_p}, \Sigma_{Z_p})$ and $\mathcal{N}(\mu_{Z_q},\Sigma_{Z_q})$ learned by the prior and recognition networks in the CVAE. The predicted poses can be recovered and outputted from the  predicted bone vectors in relative coordinates.


\section{Endpoint Loss Equations}
\label{appendix:endpoint_loss_equations}


We consider six endpoint tracks - left and right arms, legs and faces. For completeness, here is the endpoint error equation for the left arm (LA) again
\begin{align}
\label{eqn:left_arm}
\begin{split}
\mathcal{L}_{LA} &= \left \| \sum_n (\hat{J}_n - J_{parent(n)})\right \|,  n\in \left \{ 6,8,10\right \}
\\
&=  \left \| \left (\hat{J}_{10} - {J}_{8} \right ) + \left (\hat{J}_{8} - {J}_{6} \right ) + \left (\hat{J}_{6} - {J}_{18} \right ) \right \| ,
\end{split}
\end{align}
%
where $n$ denotes the joint endpoints along the left arm, the indices $\left \{ 6,8,10\right \}$ correspond to shoulder, elbow, and wrist, respectively, and 18 is the root joint. 

The endpoint error for the right arm is
\begin{align}
\label{eqn:rightarm}
\begin{split}
\mathcal{L}_{RA} &= \left \| \sum_n (\hat{J}_n - J_{parent(n)})\right \|,  n\in \left \{ 7,9,11\right \}
\\
&= \left \|  \left (\hat{J}_{11} - {J}_{9} \right ) + \left (\hat{J}_{9} - {J}_{7} \right ) + \left (\hat{J}_{7} - {J}_{18} \right ) \right \| ,
\end{split}
\end{align}
%
where $n$ denotes the joint endpoints along the right arm.

The endpoint error terms for the left and right legs are:
\begin{align}
\label{eqn:leftleg}
\begin{split}
\mathcal{L}_{LL} &= \left \| \sum_n (\hat{J}_n - J_{parent(n)})\right \|,  n\in \left \{ 12,14,16\right \}
\\
&= \left \|  \left (\hat{J}_{16} - {J}_{14} \right ) + \left (\hat{J}_{14} - {J}_{12} \right ) + \left (\hat{J}_{12} - {J}_{18} \right ) \right \| ,
\end{split}
\end{align}
%
\begin{align}
\label{eqn:rightleg}
\begin{split}
\mathcal{L}_{RL} &= \left \| \sum_n (\hat{J}_n - J_{parent(n)})\right \|,  n\in \left \{ 13,15,17\right \}
\\
&=  \left \| \left (\hat{J}_{17} - {J}_{15} \right ) + \left (\hat{J}_{15} - {J}_{13} \right ) + \left (\hat{J}_{13} - {J}_{18} \right ) \right \| ,
\end{split}
\end{align}
where $n$ denotes the joint endpoints along the left and right legs, the indices  $\left \{ 12,14,16\right \}$ and $\left \{ 13, 15,17\right \}$ correspond to left and right hip, knee, and ankle joints, respectively, and 18 is the root joint.

The endpoint error terms for the left and right face are:
\begin{align}
\label{eqn:leftface}
\begin{split}
\mathcal{L}_{LF} &= \left \| \sum_n (\hat{J}_n - J_{parent(n)})\right \|,  n\in \left \{ 1,2,4\right \}
\\
&=  \left \| \left (\hat{J}_{4} - {J}_{2} \right ) + \left (\hat{J}_{2} - {J}_{1} \right ) + \left (\hat{J}_{1} - {J}_{18} \right ) \right \| ,
\end{split}
\end{align}
%
\begin{align}
\label{eqn:rightface}
\begin{split}
\mathcal{L}_{RF} &= \left \| \sum_n (\hat{J}_n - J_{parent(n)})\right \|,  n\in \left \{ 1,3,5\right \}
\\
&=  \left \| \left (\hat{J}_{5} - {J}_{3} \right ) + \left (\hat{J}_{3} - {J}_{1} \right ) + \left (\hat{J}_{1} - {J}_{18} \right ) \right \| ,
\end{split}
\end{align}
where $n$ denotes the joint endpoints along the left and right sides of the face, the indices  $\left \{ 2, 4\right \}$ and $\left \{ 3, 5\right \}$ correspond to left and right eye and ear, respectively, 1 is the nose, and 18 is the root joint. 

The complete endpoint loss is computed as 
\begin{equation}
\label{eqn:completeendpointloss}
L_{E} = \mathcal{L}_{LA} + \mathcal{L}_{RA} + \mathcal{L}_{LF} + \mathcal{L}_{RF} + \mathcal{L}_{LL} + \mathcal{L}_{RL}.
\end{equation}
%



\section{Training and Testing Procedures}
The pedestrian anomaly detection problem is set up with a training set and a testing set, where the videos in training only contains ``normal'' events (such as walking activities), and no anomalies. Testing videos can contain both normal and abnormal/anomalous activities. The goal of our BiPOCO pipeline is to detect the anomalies in the testing set. The raw input data are in the form of 2-D RGB camera videos containing human/pedestrian activities, such as walking, running, jumping, etc. The output of our method is the detected anomaly scores for each frame in the video. 

During training, to obtain pose trajectories, we first extracted pedestrian poses in 2-D skeleton joint representation, $\mathbf{J}$, using AlphaPose pose estimator \cite{fang2017rmpe,xiu2018poseflow}. The bone vectors, $\mathbf{B}$, as described in Appendix \ref{appendix:cvae}, were computed based on $\mathbf{J}$ and were passed into BiPOCO model. The pose trajectories were broken down into subsequences of length $\tau+\delta$, where $\tau$ and $\delta$ are observation and prediction horizons, respectively. A separate model is trained for each input/output length $\tau=\delta\in \{3,5,13,25\}$ using the proposed loss function (Eq.\eqref{eqn:allloss}) with the pose constraints. 

Given testing videos, we first extract pedestrian poses using AlphaPose, similar to training. Then, the trained BiPOCO model was used for inference to predict pedestrian future poses for testing sequences. Since training data contains only ``normal'' (i.e., walking) sequences, the assumption is that the BiPOCO model learns the normal walking pattern and will predict more accurately on normal walking activities, whereas the predicted poses will differ greatly from the ground truth when testing on anomalous activities. This assumption fits the definition of ``anomaly'' that anomalous events differ from expectations. After obtaining the predicted poses for testing pedestrians, the joint error was computed and used for anomaly detection as described in Section~\ref{sec:anomaly}. The final output of the pipeline is the frame-level anomaly scores for all testing videos.

Additional details such as code, pretrained models, input JSON files (pedestrian trajectories obtained from AlphaPose), and some output PKL files (pedestrian trajectories obtained from trained model) can be found at  \url{https://github.com/akanuasiegbu/BiPOCO}.

\begin{algorithm}[t]
\caption{BiPOCO for Anomaly Detection} 
\label{alg:1}
\begin{algorithmic} 
\State  \textbf{TRAINING} 
\Require Training videos 
\State Extract pose joints $\mathbf{J}$ for all pedestrians using AlphaPose
\State Compute bone vectors $\mathbf{B}_n$ $\gets {\mathbf{J}}_{parent(n)} -{\mathbf{J}}_{n}, \forall n=\{1,...,17\}$ 
\State Train BiPOCO using  Eq.\eqref{eqn:allloss}\\
\Return Trained BiPOCO model
\Statex 
\Statex  \textbf{TESTING}
\Require Testing videos; Trained BiPOCO model
\State Extract pose joints $\mathbf{J}^{test}$ for all testing pedestrians using AlphaPose
\State Obtain predicted joints $\hat{\mathbf{J}}^{test}$ by perform inference using trained BiPOCO model
\State Compute person-level weighted joint error $e^p_{t} \gets \sum_{k=1}^{17} w_{kt} (\mathbf{J}_{kt} - \hat{\mathbf{J}}_{kt})^2$ 
\State Compute summed error and flattened error following Section C in \cite{kanu2021leveraging}
\State Compute frame-level anomaly score by max pooling
$ e^f_t \gets \max_{\forall p\in \mathcal{P}_t} e^p_{t}$
\State Compute AUC based on the anomaly scores across all frames\\
\Return Frame-level anomaly score $ e^f_t, \forall f,t$; AUC score
\end{algorithmic}
\end{algorithm}

\section{Implementation Details}
\label{appendix:implementation}


To train the bi-directional trajectory predictor in BiPOCO,  as discussed in Section~\ref{sec:bitrap}, we first transform the poses into relative coordinates by using the bone vectors $\mathbf{B}_*$, and then use a min-max normalization to normalize the pose data, where the max is the width and height of the video frames and the min is set to zero. To compute the joint loss in the pose constraints, we first un-normalize the predicted joints to obtain joints in the global (image) coordinate (in pixels), compute the joint loss, and then re-normalize it based on the video width and height. Our model input has a dimension of 36 (34 from the AlphaPose in COCO format -- 17 keypoints $\times$ 2 dimensions, plus 2 for the root joint). We also pass in the AlphaPose confidence scores in order to compute the weighted squared (prediction) error (see Section~\ref{sec:anomaly}). Three NVIDIA GPUs were used for training in our experiments, a Tesla V100 SXM2 (16 GB), a TITAN V (12 GB), and a TITAN Xp. The  number of epochs for the Avenue dataset was 250 for timescales 3 and 5, and 500 epochs for longer (13/25) timescales. For the Shanghaitech dataset, we ran for 500 epochs. A learning rate scheduler with 
$ReduceLROnPlateau$ was used to train the BiPOCO predictor.

\section{Detection Results on ShanghaiTech}
\label{appendix:st}

We present an ablation study on the compositional pose constraints on ShanghaiTech and HR-ShanghaiTech datasets,  supplementary to Table~\ref{table:frame_auc_error_all}. We report anomaly detection AUC results on timescales $\{3, 5, 13, 25\}$ for both Summed Error and Flattened Error measures. Similar to our observation on the Avenue dataset, the combination of bone and joint losses contributes to a higher AUC and adding the pose contraints in general performs better than using trajectory loss only.

\begin{table*}[h!]
\caption{Ablation study of the pose constraints on ShanghaiTech and HR-ShanghaiTech datasets: Bone(B), Endpoint(E), Joint(J), All losses(All), and Trajectory loss only(None). \textcolor{blue}{Blue}: best AUC results at each timescale. \textbf{Bold}: best AUC overall.}
\label{table:ablation_st} 
\centering
\begin{tabular}{p{2cm}|c|cccccccc} 
 \hline
  \multirow{1}{*}{Dataset}& \multirow{1}{*}{Timescale}& \multicolumn{7}{c}{Loss Terms} \\\cline{3-10}
  \multirow{4}{1.3cm}{ \hspace{15mm} HR-ShanghaiTech - SE} &    & B & E &J & B-E& B-J& E-J& All & None\\ \hline
  &$\tau=\delta=3$   & \textcolor{blue}{0.724}	&	0.625	&	0.720	&	0.683	&	0.720	&	0.723	&	\textcolor{blue}{0.724}  &0.715\\
  & $\tau=\delta=5$ & 0.740	&	0.680	&	\textcolor{blue}{0.743}	&	0.728	&	0.742	&	0.740	&	0.740	&0.736	\\
  & $\tau=\delta=13$ & 0.547	&	0.728	&	0.725	&	0.734	&\textcolor{blue}{0.735}	&	0.728	&	0.720 & 0.729\\
  & $\tau=\delta=25$ & 0.652	&	\textcolor{blue}{0.671}	&	0.525	&	0.668	&	0.665	&	0.665	&	0.669 & 0.520\\\hline
  \multirow{4}{1.3cm}{ \hspace{15mm} HR-ShanghaiTech - FE} &$\tau=\delta=3$  & \textcolor{blue}{0.724}	&	0.627	&	0.722	&	0.676	&	0.715	&	0.722	&	0.722 &0.717 \\	
  & $\tau=\delta=5$ & 0.743	&	0.687	&	0.745	&	0.733	&	\textbf{\textcolor{blue}{0.749}}	&	0.747	&	0.742 & 0.739\\
  & $\tau=\delta=13$ & 0.555	&	0.740	&	0.733	&	0.739	&	0.743	&	\textcolor{blue}{0.747}	&	0.734 &0.744\\
  & $\tau=\delta=25$ & 0.658	&	0.683	&	0.556	&	\textcolor{blue}{0.690}	&	0.675	&	0.668	&	0.676	& 0.543\\	\hline
 \multirow{4}{1.3cm}{ \hspace{15mm} ST - SE} &$\tau=\delta=3$ & \textcolor{blue}{0.709}	&	0.620	&	0.706	&	0.672	&	0.705	&	\textcolor{blue}{0.709}	&	\textcolor{blue}{0.709}	& 0.700\\
  & $\tau=\delta=5$ &  0.727	&	0.671	&	\textcolor{blue}{0.730}	&	0.716	&	0.729	&	0.726	&	0.729 & 0.724\\
  & $\tau=\delta=13$ & 0.552	&	0.717	&	0.714	&	0.722	&	\textcolor{blue}{0.724}	&	0.716	&	0.709& 0.717 \\
  & $\tau=\delta=25$ &  0.646	&	\textcolor{blue}{0.664}	&	0.523	&	0.661	&	0.657	&	0.656	&	0.661	&	0.522\\\hline 
  \multirow{4}{1.1cm}{ \hspace{15mm} ST - FE}& $\tau=\delta=3$ & \textcolor{blue}{0.709}	&	0.621	&	0.708	&	0.665	&	0.700	&	0.707	&	0.708 & 0.703 \\
  & $\tau=\delta=5$ & 0.729	&	0.678	&	0.731	&	0.721	&	\textcolor{blue}{0.735}	&	0.734	&	0.729 &0.726\\
  & $\tau=\delta=13$ & 0.548	&	0.730	&	0.723	&	0.729	&	0.734	&	\textbf{\textcolor{blue}{0.737}}	&	0.725& 	0.734\\
  & $\tau=\delta=25$ & 0.655	&	0.674	&	0.553	&	\textcolor{blue}{0.681}	&	0.670	&	0.662	&	0.667	& 0.549\\ \hline
  \end{tabular}
\end{table*}

\begin{table*}[h!]
\caption{Joint Error results for normal/anomalous instances on Avenue and HR-Avenue datasets. 	\textcolor{blue}{Blue} corresponds to best AUC values from Summed Error for the different timescales. \textbf{Bold} corresponds to the largest difference between the normal and anomalous joint errors.}
\label{table:Joint Error} 
\centering
\resizebox{\textwidth}{!}{
\begin{tabular}{c|c|cccccccc} 
 \hline
  \multirow{1}{*}{Dataset}& \multirow{1}{*}{Timescale}& \multicolumn{7}{c}{Loss Terms} \\\cline{3-10}
  \multirow{4}{*}{ HR-Avenue} &    & B & E &J & B-E& B-J& E-J& All & None\\ \hline
  &$\tau=\delta=3$    & 2.52/2.95 & \textbf{	\textcolor{blue}{4.84/6.68}} &2.29/2.81 & 2.77/3.52 & 2.45/2.89 &2.39/2.85 &2.54/2.84  & 2.38/2.88 \\
  & $\tau=\delta=5$ & 2.89/3.46 &3.92/4.92 & 2.73/3.68 & 	\textcolor{blue}{2.78/3.70} &\textbf{2.53/3.65} &2.77/3.33 &3.01/3.65 & 2.71/3.52 \\
  & $\tau=\delta=13$ & 3.36/4.97 & 4.00/5.05& 	\textcolor{blue}{3.36/4.79} & \textbf{3.28/5.10} & 3.23/4.90 & 3.33/4.94 & 4.24/4.80 & 3.82/5.28 \\
  & $\tau=\delta=25$ & 4.68/5.82 & 3.98/5.96 & 4.40/5.88 & 4.36/5.98 & 	\textbf{\textcolor{blue}{4.35/6.60}} & 4.94/6.72 & 4.62/6.39 & 4.31/6.09 \\ \hline
  \multirow{4}{*}{Avenue}& $\tau=\delta=3$ & 2.53/3.30 & \textbf{	\textcolor{blue}{4.83/6.78}} & 2.30/3.27 &2.77/3.92 &2.46/3.30 & 2.40/3.30 &2.56/3.31 & 2.39/3.34\\
  & $\tau=\delta=5$ & 2.91/3.84  & 	\textcolor{blue}{3.93/5.24} & \textbf{2.74/4.09} & 2.79/3.97 & 2,53/3.89 & 2.78/3.70 & 3.03/4.11 & 2.72/3.94\\
  & $\tau=\delta=13$ & 3.37/5.16 & 4.05/5.33 & 3.37/5.27 & \textbf{	\textcolor{blue}{3.29/5.36}} & 	\textcolor{blue}{3.23/5.13} & 3.34/5.24 &4.27/5.33 & 3.87/5.60 \\
  & $\tau=\delta=25$ & 4.70/6.29 & 3.99/6.33 & 4.45/6.28 & 4.37/6.33 & \textbf{	\textcolor{blue}{4.34/6.71}} & 4.93/6.94 & 4.61/6.50 & 4.29/6.17\\ \hline
  \end{tabular}}
\end{table*}

\section{Joint Error Results}
\label{appendix:joint_error}

We present joint error results to study the prediction performance of our BiPOCO pipeline. The joint error is computed as the root mean squared error between predicted and ground truth joints across all prediction timesteps
\begin{equation}
\label{eqn:jointerror}
e_j = \frac{1}{\delta*17} \sum_{t=1}^{\delta} \sqrt{ \sum_{k=1}^{17} (\hat{\mathbf{J}}_{kt} - \mathbf{J}_{kt})^2 }.
\end{equation}

Table~\ref{table:Joint Error} presents joint error results for normal and anomalous instances on Avenue and HR-Avenue datasets. We find that our best performing model with the highest AUC (using the endpoint loss, timescale 3) has higher joint prediction error for both normal and anomalous frames (4.84/6.68 and 4.83/6.78) compared to other loss terms, but the difference between anomalous and normal joint errors for using the endpoint loss is also the largest. This corresponds to a high detection AUC performance as shown in Table~\ref{table:frame_auc_error_all}. This implies that, for timescale 3, using the other loss terms (bone and joints) provided a more accurate prediction on both normal and anomalous instances, which made it harder to separate normal and anomalous instances. On the other hand, the endpoint loss, with emphasis on constraining the extremities, learned the ``normal'' patterns better than the untrained, anomalous events, and, thus, resulted in a larger difference between the normal and anomalous cases and can, thus, better distinguish between the two, leading to a higher AUC detection performance. 

\section{Visual Results of Pose Prediction}

\begin{figure}[ht!]
    \centering
    \resizebox{.95\textwidth}{!}{
    \begin{minipage}{0.33\linewidth}
        \centering
        \includegraphics[width=2.2in]{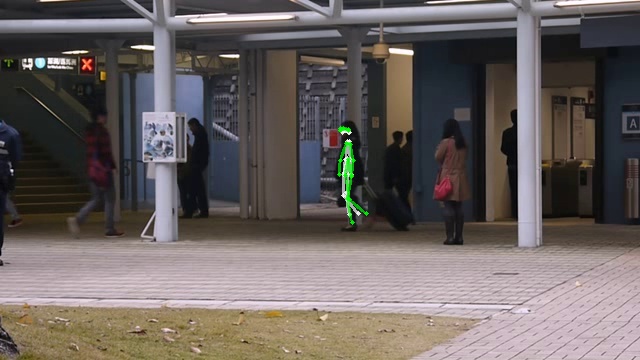}\\
        195
    \end{minipage}
    \begin{minipage}{0.33\linewidth}
        \centering
        \includegraphics[width=2.2in]{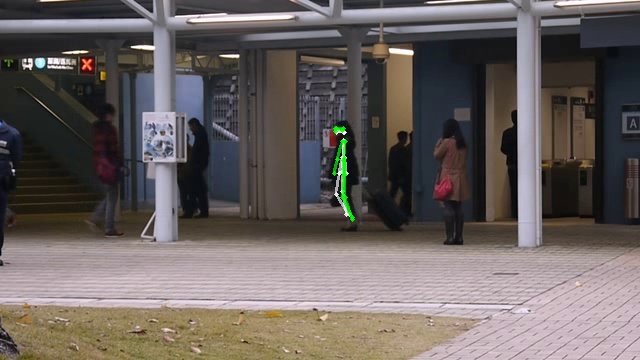}\\
        197
    \end{minipage}
    \begin{minipage}{0.33\linewidth}
        \centering
        \includegraphics[width=2.2in]{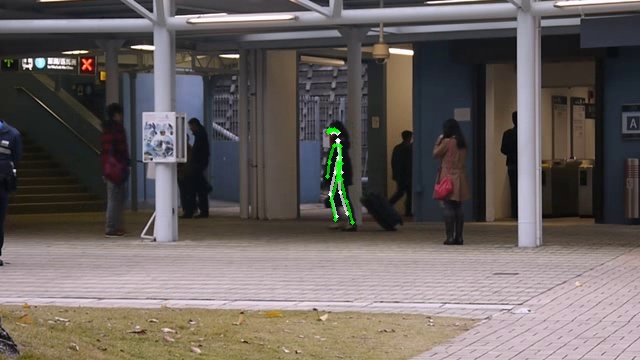}\\
        199
    \end{minipage} }
    
    \resizebox{.95\textwidth}{!}{
     \begin{minipage}{0.33\linewidth}
        \centering
        \includegraphics[width=2.2in]{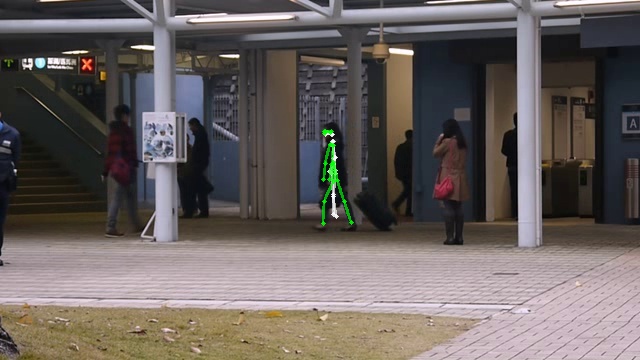}\\
        201
    \end{minipage}
    \begin{minipage}{0.33\linewidth}
        \centering
        \includegraphics[width=2.2in]{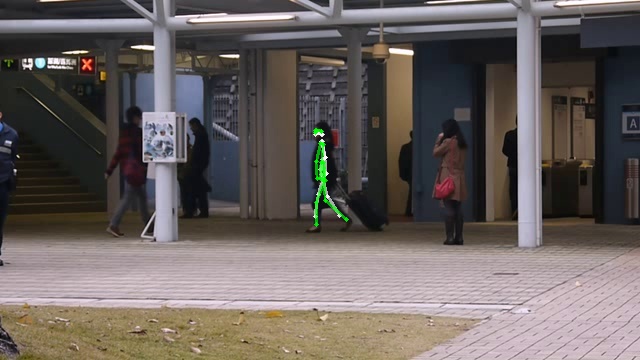}\\
        204
    \end{minipage}
    \begin{minipage}{0.33\linewidth}
        \centering
        \includegraphics[width=2.2in]{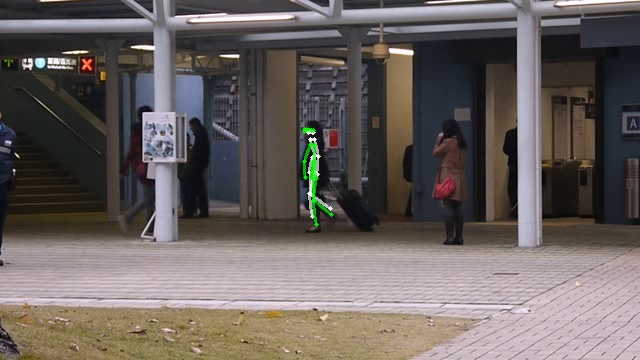}\\
        207
    \end{minipage}}
    \vspace{-3mm}
    \caption{Visualization of ground truth (green) and predicted poses (white) for a ``normal'' pedestrian walking sequence (timescale 13, Avenue dataset video 05 frames \#195 to \#207, pedestrian ID 05).  Best viewed in color. The numbers under the images are frame numbers.}
    \label{fig:vis1}
\end{figure}

\begin{figure}[ht!]
    \centering
    \resizebox{.95\textwidth}{!}{
    \begin{minipage}{0.33\linewidth}
        \centering
        \includegraphics[width=2.2in]{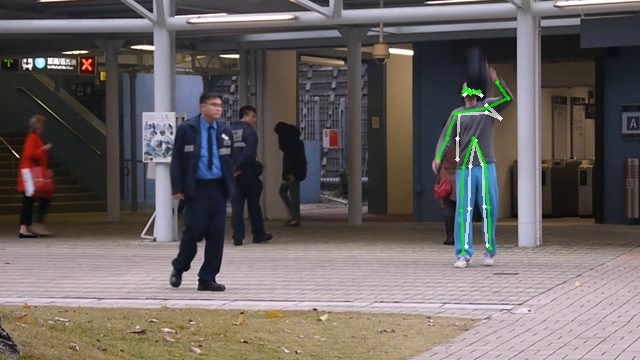}\\
        476
    \end{minipage}
    \begin{minipage}{0.33\linewidth}
        \centering
        \includegraphics[width=2.2in]{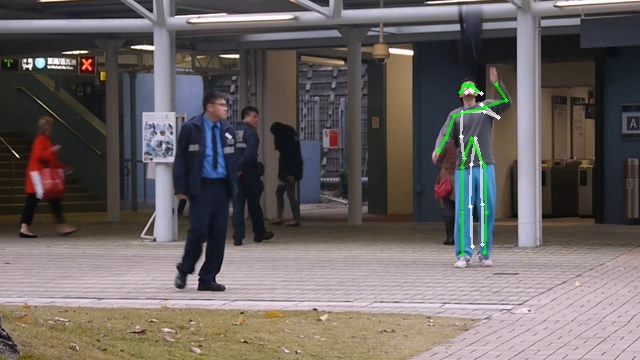}\\
        478
    \end{minipage}
    \begin{minipage}{0.33\linewidth}
        \centering
        \includegraphics[width=2.2in]{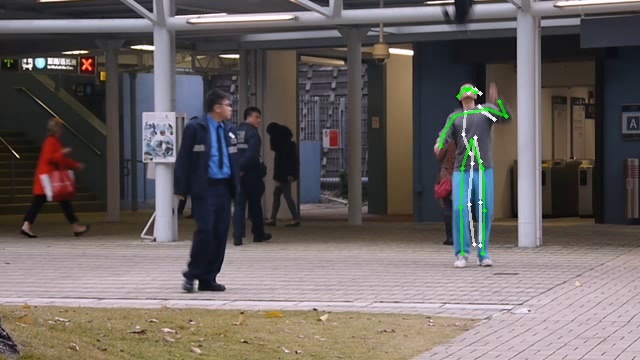}\\
        480
    \end{minipage}} 
    
    \resizebox{.95\textwidth}{!}{
     \begin{minipage}{0.33\linewidth}
        \centering
        \includegraphics[width=2.2in]{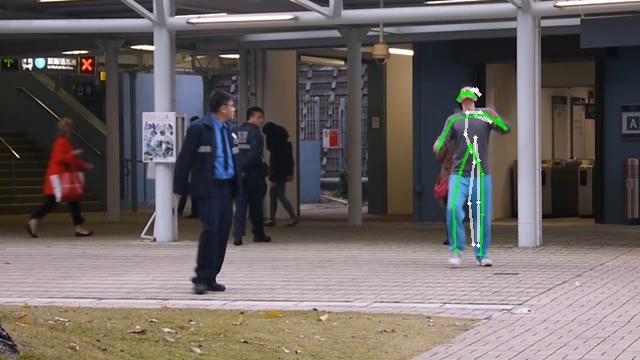}\\
        482
    \end{minipage}
    \begin{minipage}{0.33\linewidth}
        \centering
        \includegraphics[width=2.2in]{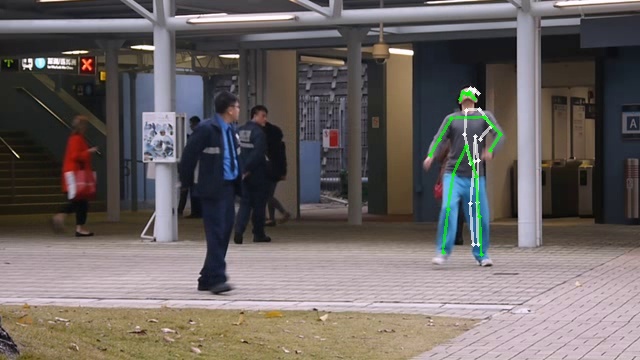}\\
        485
    \end{minipage}
    \begin{minipage}{0.33\linewidth}
        \centering
        \includegraphics[width=2.2in]{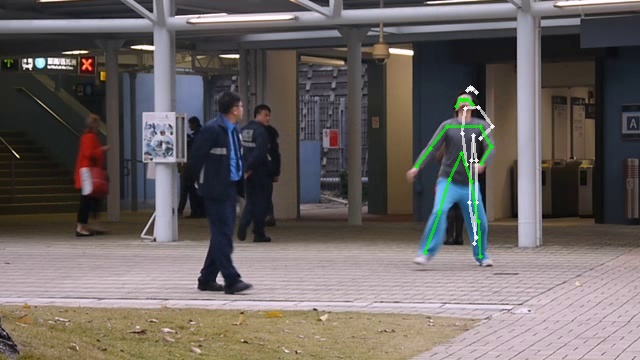}\\
        488
    \end{minipage}}
    \vspace{-2mm}
    \caption{Visualization of ground truth (green) and predicted poses (white) for an anomalous activity, where the pedestrian is jumping up and also throwing a bag (timescale 13, Avenue dataset video 05 frames \#467 to \#488, pedestrian ID 06).  Best viewed in color. The numbers under the images are frame numbers.}
    \label{fig:vis2}
\end{figure}

We present two visualization examples of the BiPOCO pose prediction results. Figure~\ref{fig:vis1} presents visual results of a ``normal'' walking pedestrian sequence in the Avenue dataset. Figure~\ref{fig:vis2} presents visual results of an anomalous event where a pedestrian is throwing a bag. As shown, the predicted pose (white) for the walking pedestrian overlaps with the ground truth pose (green) well, and captures the movements of the legs during walking. For the ``throwing'' sequence, since BiPOCO was trained on normal walking sequences only, the predicted pose for throwing a bag did not quite match the ground truth, particularly for the actions in the left arm, which is as expected. It is also worth noting that BiPOCO can produce visually physically plausible poses for both normal and anomalous activities.

\vspace{-10mm}
\end{document}